\documentclass[5p,final,authoryear,twocolumn]{elsarticle}
\usepackage{graphicx}
\usepackage{amssymb}

\usepackage{array}
\usepackage{amsmath}
\usepackage{hyperref}
\usepackage[pdftex,table]{xcolor}
\usepackage{booktabs}

\newcommand{\otoprule}{\midrule[\heavyrulewidth]}

\journal{Medical Image Analysis}

\begin{document}

\begin{frontmatter}


\title{`Squeeze \& Excite' Guided Few-Shot Segmentation of Volumetric Images}

\makeatletter
\def\@author#1{\g@addto@macro\elsauthors{\normalsize%
    \def\baselinestretch{1}%
    \upshape\authorsep#1\unskip\textsuperscript{%
      \ifx\@fnmark\@empty\else\unskip\sep\@fnmark\let\sep=,\fi
      \ifx\@corref\@empty\else\unskip\sep\@corref\let\sep=,\fi
      }%
    \def\authorsep{\unskip,\space}%
    \global\let\@fnmark\@empty
    \global\let\@corref\@empty  
    \global\let\sep\@empty}%
    \@eadauthor={#1}
}
\makeatother


\author{Abhijit Guha Roy$^{a,b}$ \corref{cor1}} 
\cortext[cor1]{{A. Guha Roy, S. Siddiqui and S. P\"olsterl has contributed equally to this work. Corresponding Author Address: KJP, LMU, Waltherstr. 23, 80337 M\"unchen, Germany; Email: abhi4ssj@gmail.com}}

\author{Shayan Siddiqui$^{a,b}$ \corref{cor1}}
\author{Sebastian P\"{o}lsterl$^{a}$ \corref{cor1}}
\author{Nassir Navab$^{b,c}$} 
\author{Christian Wachinger$^{a}$}

\address{$^{a}$Artificial Intelligence in Medical Imaging (AI-Med), Department of Child and Adolescent Psychiatry, LMU M\"{u}nchen, Germany. \\
$^{b}$Computer Aided Medical Procedures, Department of Informatics, Technical University of Munich, Germany. \\
$^{c}$Computer Aided Medical Procedures, Johns Hopkins University, Baltimore, USA.
}

\begin{abstract}
Deep neural networks enable highly accurate image segmentation, but require large amounts of manually annotated data for supervised training. Few-shot learning aims to address this shortcoming by learning a new class from a few annotated support examples. We introduce, a novel few-shot framework, for the segmentation of volumetric medical images with only a few annotated slices. Compared to other related works in computer vision, the major challenges are the absence of pre-trained networks and the volumetric nature of medical scans. We address these challenges by proposing a new architecture for few-shot segmentation that incorporates `squeeze \& excite' blocks. 
Our two-armed  architecture  consists  of  a  conditioner  arm,  which  processes  the  annotated  support  input  and  generates  a task-specific representation.
This representation is passed on to the segmenter arm that uses this information to segment the new query image. To facilitate efficient interaction between the conditioner and the segmenter arm, we propose to use `channel squeeze \& spatial excitation' blocks -- a light-weight computational module -- that enables heavy interaction between both the arms with negligible increase in model complexity. This contribution allows us to perform image segmentation without relying on a pre-trained model, which generally is unavailable for medical scans. Furthermore, we propose an efficient strategy for volumetric segmentation by optimally pairing a few slices of the support volume to all the slices of the query volume. We perform experiments for organ segmentation on whole-body contrast-enhanced CT scans from the Visceral Dataset. Our proposed model outperforms multiple baselines and existing approaches with respect to the segmentation accuracy by a significant margin. The source code is available at \href{https://github.com/abhi4ssj/few-shot-segmentation}{\texttt{https://github.com/abhi4ssj/few-shot-segmentation}}.
\end{abstract}


\end{frontmatter}


\section{Introduction}
\label{sec:intro}

Fully convolutional neural networks (F-CNNs) have achieved state-of-the-art performance in semantic image segmentation for both natural~\citep{jegou2017one, zhao2017pyramid, long2015fully, noh2015learning} and medical images~\citep{ronneberger2015u, milletari2016v}.
Despite their tremendous success in image segmentation, they are of limited use when only a few labeled images are available.
F-CNNs are in general highly complex models with millions of trainable weight parameters that require thousands of densely annotated images for training to be effective.
A better strategy could be to adapt an already trained F-CNN model to segment a new semantic class from a few labeled images.
This strategy often works well in computer vision applications where a pre-trained model is used to provide a good initialization and is subsequently fine-tuned with the new data to tailor it to the new semantic class.
However, fine-tuning an existing pre-trained network without risking over-fitting still requires a fair amount of annotated images (at least in the order of hundreds). When dealing in an extremely low data regime, where only a single or a few annotated images of the new class are available, such fine-tuning based transfer learning often fails and may cause overfitting~\citep{shaban2017one, rakelly2018few}.

\begin{figure*}[t]
\center{\includegraphics[width=0.85\textwidth]{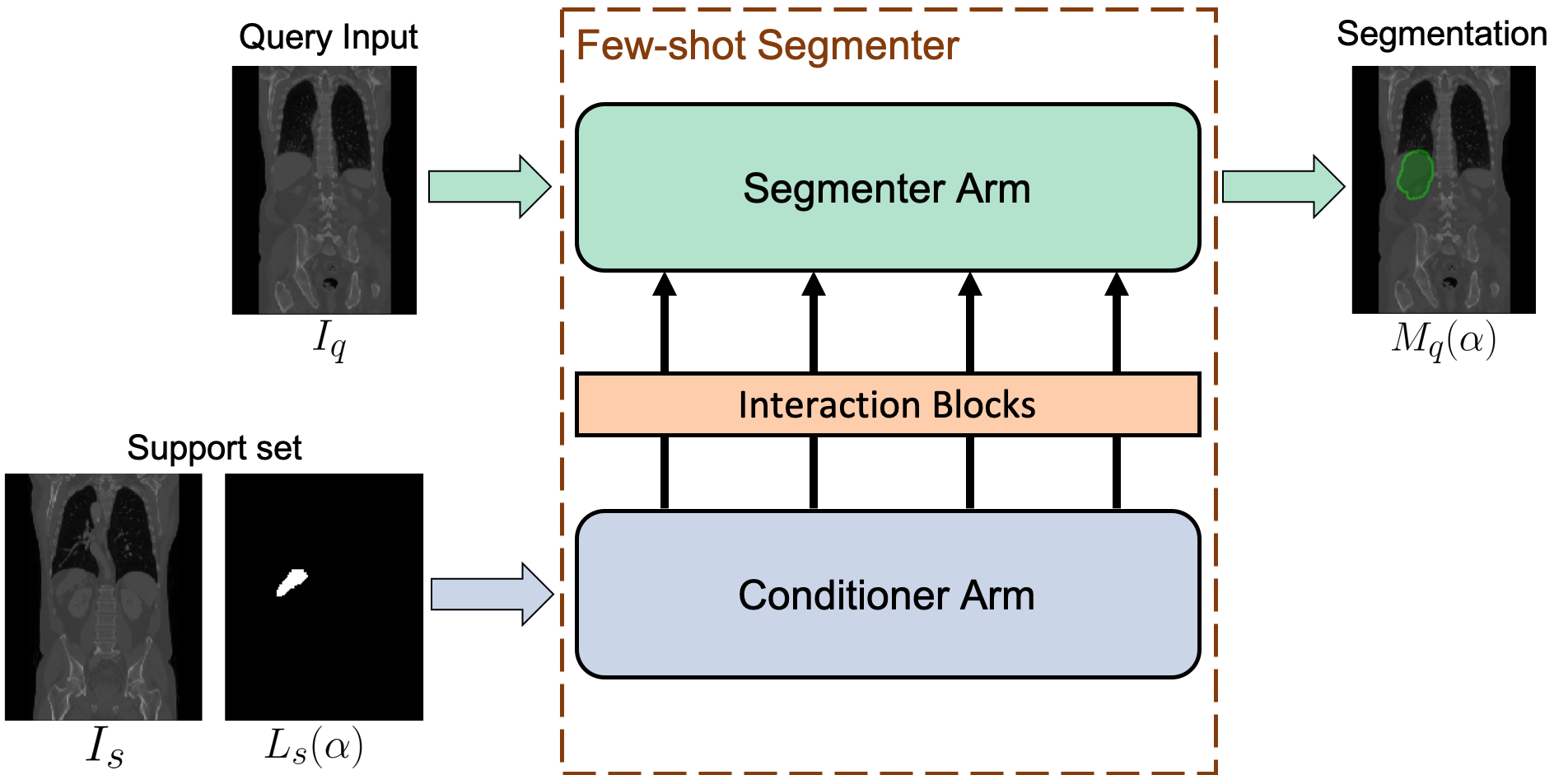}}
\caption{Overview of the few-shot segmentation framework. The support set consists of an image slice $I_s$ and the corresponding annotation for the new semantic class $L_s(\alpha)$ (here $\alpha$ is the class liver). We pass the support set through the conditioner arm, whose information is conveyed to the segmenter arm via interaction blocks. The segmenter arm uses this information and segments a query input image $I_q$ for the class $\alpha$ generating the label map $M_q(\alpha)$. Except for the support set, the few-shot segmenter has never seen annotations of a liver before.}
\label{fig:graphAbs}
\end{figure*}

Few-shot learning is a machine learning technique that aims to address situations where an existing model needs to generalize to an unknown semantic class with a few examples at a rapid pace~\citep{fei2006one, miller2000learning, fei2006knowledge}. The basic concept of few-shot learning is motivated by the learning process of humans, where learning new semantics is done rapidly with very few observations, leveraging strong prior knowledge acquired from past experience. 
While few-shot learning for image classification and object detection is a well studied topic, few-shot learning for semantic image segmentation with neural networks has only recently been proposed~\citep{shaban2017one,rakelly2018few}. It is an immensely challenging task to make dense pixel-level high-dimensional predictions in such an extremely low data regime.
But at the same time, few-shot learning could have a big impact on medical image analysis because it addresses learning from scarcely annotated data, which is the norm due to the dependence on medical experts for carrying out manual labeling. 
In this article, we propose a few-shot segmentation framework designed exclusively for segmenting volumetric medical scans.
A key to achieve this goal is to integrate the recently proposed `squeeze \& excite' blocks within the design of our novel few-shot architecture~\citep{roy2018recalibrating}. 

\subsection{Background on Few-Shot Segmentation}
\label{sec:backgnd}
Few-shot learning algorithms try to generalize a model  to a new, previously unseen class with only a few labeled examples by utilizing the previously acquired knowledge from differently labeled training data. 
Fig.~\ref{fig:graphAbs} illustrates the overall setup, where we want to segment the liver in a new scan given the annotation of liver in only a single slice. 
A few-shot segmentation network architecture~\citep{shaban2017one, rakelly2018few} commonly consists of three parts: (i) a conditioner arm, (ii) a set of interaction blocks, and (iii) a segmentation arm. 
During inference, the model is provided with a support set $(I_s,L_s(\alpha))$, consisting of an image~$I_s$ with the new semantic class (or organ) $\alpha$ outlined as a binary mask indicated as~$L_s(\alpha)$. 
In addition, a query image~$I_q$ is provided, where the new semantic class is to be segmented.
The conditioner takes in the support set and performs a forward pass. This generates multiple feature maps of the support set in all the intermediate layers of the conditioner arm. This set of feature maps is referred to as \emph{task representation} as they encode the information required to segment the new semantic class. The \emph{task representation} is taken up by the interaction blocks, whose role is to pass the relevant information to the segmentation arm. The segmentation arm takes the query image as input, leverages the task information as provided by the interaction blocks and generates a segmentation mask $M_q$ for the query input~$I_q$. Thus, interaction blocks pass the information from the conditioner to the segmenter and form the backbone for few-shot semantic image segmentation. 
Existing approaches use weak interactions with a single connection either at the bottleneck or the last layer of the network~\citep{shaban2017one, rakelly2018few}.


\subsection{Challenges for Medical Few-Shot Segmentation}
\label{sec:challenges}
Existing work in computer vision on few-shot segmentation processes 2D RGB images and uses a pre-trained model for both segmenter and conditioner arm to aid  training~\citep{shaban2017one, rakelly2018few}.
Pre-trained models provide a strong prior knowledge with more powerful features from the start of training. Hence, weak interaction between  conditioner and segmenter is sufficient to train the model effectively.
The direct extension to medical images is challenging due to the lack of pre-trained models. 
Instead, both the conditioner and the segmenter need to be trained from scratch.
However, training the network in the absence of pre-trained models with weak interaction is prone to instability and mode collapse. 

Instead of weak interaction, we propose a strong interaction at multiple locations between both the arms. 
The strong interaction facilitates effective gradient flow across the two arms, which eases the training of both the arms without the need for any pre-trained model.
For effectuating the interaction, we propose our recently introduced `channel squeeze \& spatial excitation' (sSE) module~\citep{roy2018concurrent, roy2018recalibrating}. 
In our previous works, we used the sSE blocks for adaptive self re-calibration of feature maps to aid segmentation in a single segmentation network. 
Here, we use the sSE blocks to communicate between the two arms of the few-shot segmentation network. 
The block takes as input the learned conditioner feature map and  performs `channel squeeze' to learn a spatial map. This is used to perform `spatial excitation' on the segmenter feature map. We use sSE blocks between all the encoder, bottleneck and decoder blocks. 
SE blocks are well suited for effectuating the interaction between arms, as they are light-weight and therefore only marginally increase the model complexity. Despite its light-weight nature, they can have a strong impact on the segmenter's features via re-calibration.


Existing work on few-shot segmentation focused on 2D images, while we are dealing with volumetric medical scans. 
Manually annotating organs on all slices in 3D images is time consuming. 
Following the idea of few-shot learning, the annotation should rather happen on a few sparsely selected slices. 
To this end, we propose a volumetric segmentation strategy by properly pairing a few annotated slices of the support volume with all the slices of the query volume, maintaining inter-slice consistency of the segmentation.

\subsection{Contributions}
In this work, we propose:
\begin{enumerate}
    \item A novel few-shot segmentation framework for volumetric medical scans.
    \item Strong interactions at multiple locations between the conditioner and segmenter arms, instead of only one interaction at the final layer. 
    \item `Squeeze \& Excitation' modules for effectuating the interaction.
    \item Stable training from scratch without requiring a pre-trained model.
    \item A volumetric segmentation strategy that optimally pairs the slices of query and support volumes.
\end{enumerate}

\subsection{Overview}
We discuss related work in Sec.~\ref{sec:priorArt}, present our few-shot segmentation algorithm in Sec.~\ref{sec:method}, the experimental setup in Sec.~\ref{sec:exp_setup} and experimental results and discussion in Sec.~\ref{sec:results}. We conclude with a summary of our contributions in Sec.~\ref{sec:conc}.

\section{Prior Work}
\label{sec:priorArt}


\subsection{Few-Shot Learning}
Methods for few-shot learning can be broadly divided into three groups.
The first group of methods adapts a base classifier to the new class~\citep{bart2005cross, fei2006one, hariharan2017low}. These approaches are often prone to overfitting as they attempt to fit a complex model on a few new samples. 
Methods in the second group aim to predict classifiers close to the base classifier to prevent overfitting. The basic idea is to use a two-branch network, where the first branch predicts a set of dynamic parameters, which are used by the second branch to generate a prediction~\citep{bertinetto2016learning, wang2016learning}. The third group contains algorithms that use metric learning. They try to map the data to an embedding space, where dissimilar samples are mapped far apart and similar samples are mapped close to each other, forming clusters. Standard approaches rely on Siamese architectures for this purpose~\citep{koch2015siamese, vinyals2016matching}.

\subsection{Few-Shot Segmentation using Deep Learning}
Few-shot image segmentation with deep neural networks has been explored only recently. 
In one of the earliest work, ~\citet{caelles2017one} leverage the idea of fine-tuning a pre-trained model with limited data. The authors perform video segmentation, given the annotation of the first frame. Although their model performed adequately in this application, such approaches are prone to overfitting and adapting a new class requires retraining, which hampers the speed of adaptation. \citet{shaban2017one} use a 2-arm architecture, where the first arm looks at the new sample along with its label to regress the classification weights for the second arm, which takes in a query image and generates its segmentation. \citet{dong2018few} extended this work to handle multiple unknown classes at the same time to perform multi-class segmentation. \citet{rakelly2018few} took it to an extremely difficult situation where supervision of the support set is provided only at a few selected landmarks for foreground and background, instead of a densely annotated binary mask.
Existing approaches for few-shot segmentation were evaluated on the PASCAL VOC computer vision benchmark~\citep{shaban2017one, rakelly2018few}.
They reported low segmentation scores (mean intersection over union around $40\%$), confirming that
few-shot segmentation is a very challenging task.

All of the above mentioned papers depend on pre-trained models to start the training process. Although access to pre-trained models is relatively easy for computer vision applications, no pre-trained models are available for medical imaging applications. Moreover, they use 2D RGB images, whereas we deal with 3D volumetric medical scans. This is more challenging, because there is no established strategy to select and pair support slices with the query volume. This can lead to situations where the query slice can be very different from the support slice or may not even contain the target class at all.

In the domain of medical image segmentation, recently \citet{zhao2019data} et. al. used a learnt transformation to highly augment a single annotated volume for one-shot segmentation. This differs from our approach in two aspects: (i) they use a single fully annotated volume, whereas we use annotations of only a few slices, (ii) they use a learnt representation to highly augment the single annotated volume for segmentation, whereas we use separate dataset with annotations provided for other classes.
We follow the experimental setting defined in computer vision PASCAL VOC benchmarks by~\cite{shaban2017one}.


\section{Method}
\label{sec:method}
In this section, we first introduce the problem setup, then detail the architecture of our network and the training strategy, and finally describe the evaluation strategy for segmenting volumetric scans.

\subsection{Problem Setup for Few-shot Segmentation}
The training data for few-shot segmentation $\mathcal{D}_\text{Train} = \{ (I_T^i, L_T^i(\alpha)) \}_{i=1}^N$ comprises $N$ pairs of input image $I_T$ and its corresponding binary label map $L_T(\alpha)$ with respect to the semantic class (or organ) $\alpha$. All the semantic classes $\alpha$ which are present in the label map $L_T^i \in \mathcal{D}_\text{Train}$ belong to the set $\mathcal{L}_\text{Train} = \{ 1, 2, \dots, \kappa \}$, i.e., $\alpha \in \mathcal{L}_\text{Train}$. Here $\kappa$ indicates the number of classes (organs) annotated in the training set.
The objective is to learn a model $\mathcal{F}(\cdot)$ from $\mathcal{D}_\text{Train}$, such that given a support set $(I_s, L_s(\hat{\alpha})) \notin \mathcal{D}_\text{Train}$ for a new semantic class $\hat{\alpha}\in\mathcal{L}_\text{Test}$ and a query image $I_q$, the binary segmentation $M_q(\hat{\alpha})$ of the query is inferred. 
Fig.~\ref{fig:graphAbs} illustrates the setup for the test class $\hat{\alpha} = \text{liver}$ in a CT scan. The semantic classes  for  training and testing are mutually exclusive, i.e., $\mathcal{L}_\text{Train} \cap \mathcal{L}_\text{Test} = \emptyset$.

One fundamental difference of few-shot segmentation to few-shot classification or object detection is that test classes $\mathcal{L}_\text{Test}$  might already appear in the training data as the background class.
For instance, the network has already seen the liver on many coronal CT slices as part of the background class, although liver was not a part of the training classes. 
This potentially forms prior knowledge that could be utilized during testing, when only a few examples are provided with the liver annotated.

\begin{figure*}[h]
\center{\includegraphics[width=\textwidth]{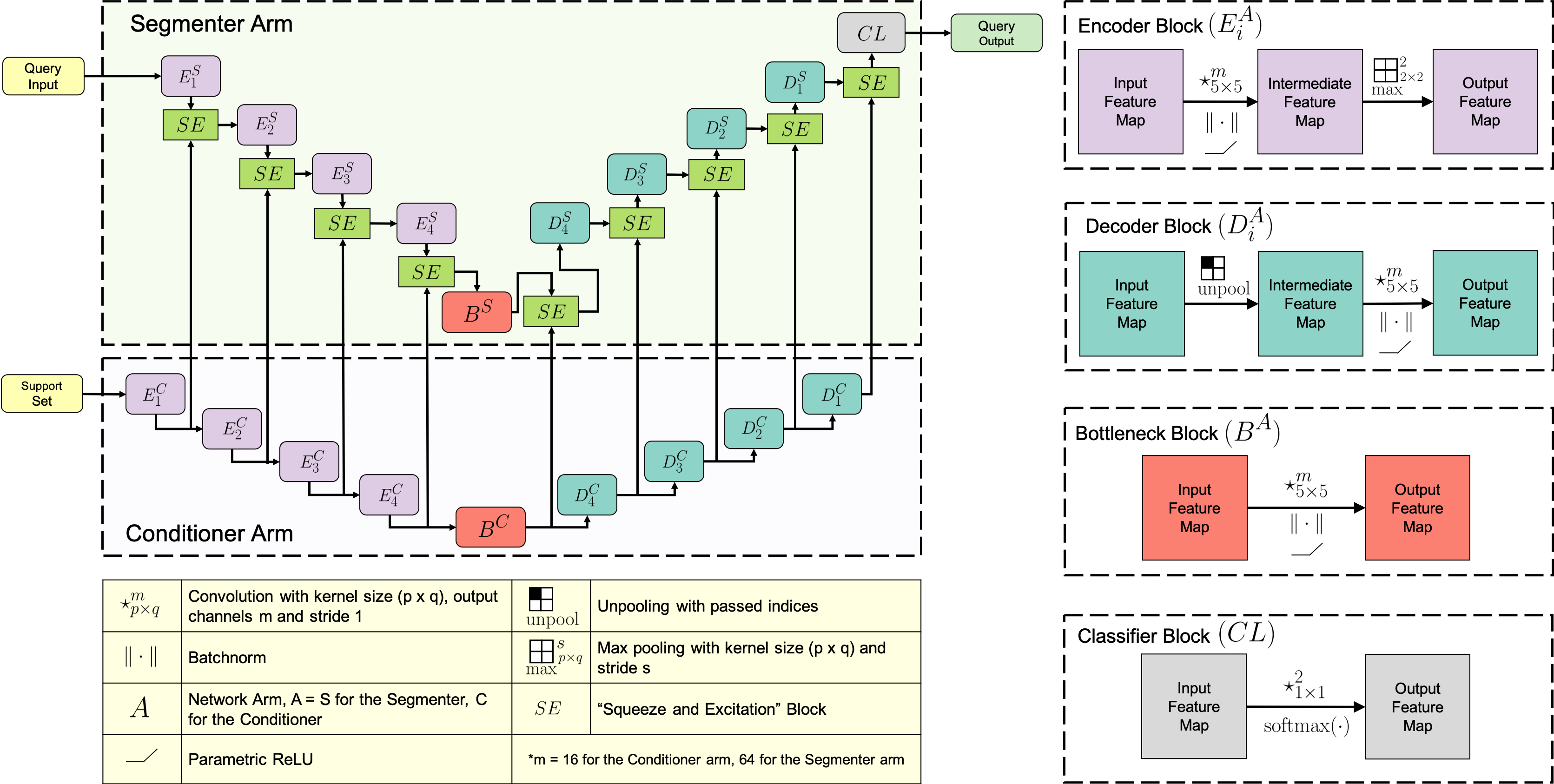}}
\caption{Illustration of the architecture of the few-shot segmenter. To the left, we show a block diagram with arrows illustrating the encoder-decoder based conditioner arm (bottom) and segmenter arm (top). Interaction between them is shown by SE blocks, which is detailed in Fig.~\ref{fig:squeeze}. To the right, the operational details of the encoder block, decoder block, bottleneck block and the classifier block are provided.}
\label{fig:architecture}
\end{figure*}

\subsection{Architectural Design}
\begin{figure}[h]
\center{\includegraphics[width=0.47\textwidth]{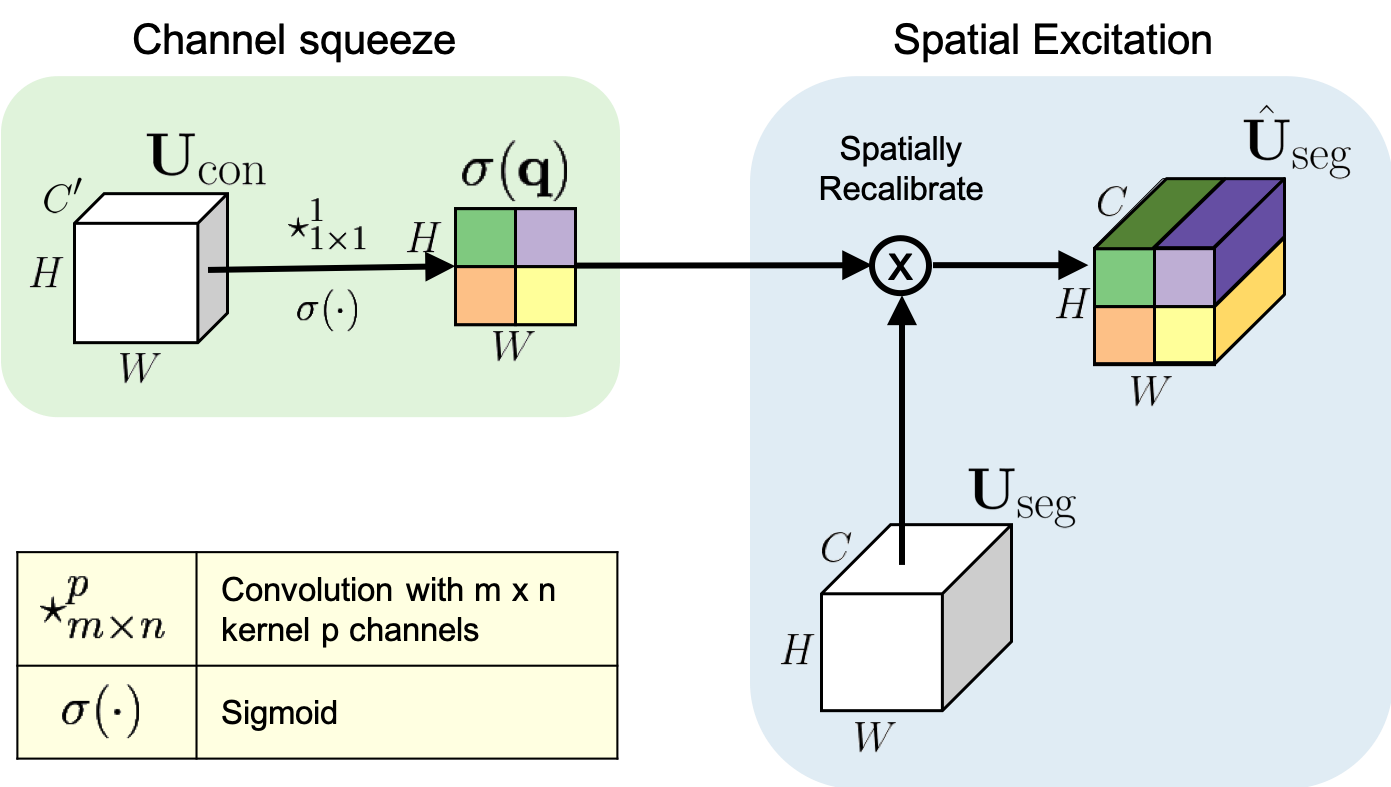}}
\caption{Illustration of the architecture of the `channel squeeze \& spatial excitation' (sSE) module, which is used as the interaction block within the few-shot segmenter. The block takes a conditioner feature map $\mathbf{U}_{\mathrm{con}}$ and a segmenter feature map $\mathbf{U}_{\mathrm{seg}}$ as inputs. `Channel squeeze' is performed on $\mathbf{U}_{\mathrm{con}}$ to generate a spatial map $\sigma(\mathbf{q})$, which is used for `spatial excitation' of $\mathbf{U}_{\mathrm{seg}}$, which promotes the interaction.}
\label{fig:squeeze}
\end{figure}

As mentioned earlier, our network architecture consists of three building blocks: (i) a conditioner arm, (ii) interaction blocks with sSE modules, and (iii) a segmenter arm. 
The conditioner arm processes the support set to model how a new semantic class (organ) looks like in an image. It efficiently conveys the information to the segmenter arm through the interaction blocks. The segmenter arm segments the new semantic class in a new query image by utilizing the information provided by the interaction blocks. 
Figs.~\ref{fig:architecture} and~\ref{fig:squeeze} illustrate the architecture in further detail, which is also described below. 

In our framework, we choose the segmenter and conditioner to have a symmetric layout, i.e., both have four encoder and decoder blocks separated by a bottleneck block. 
The symmetric layout helps in having a strong interaction between matching blocks, as feature maps  have the same spatial resolution. 
In existing approaches, conditioner and segmenter only interact via the final layer, before generating segmentation maps~\citep{shaban2017one,rakelly2018few}. 
Such weak interaction at a single location was  sufficient for their application, because they were able to use a pre-trained model, which already provides reasonably good features.
As we do not have a pre-trained network, we propose to establish a strong interaction by incorporating the sSE blocks at multiple locations. Such interactions facilitate  training the model from scratch. 

\subsubsection{Conditioner Arm}
The task of the conditioner arm is to process the support set by fusing the visual information of the support image $I_s$ with the annotation $L_s$, and generate task-specific feature maps, capable of capturing what should be segmented in the query image $I_q$. We refer to the intermediate feature maps of the conditioner as \emph{task representation}.
We provide a 2-channel input to the conditioner arm by stacking $I_s$ and binary map $L_s(\alpha)$. This is in contrast to~\citet{shaban2017one}, where they multiplied $I_s$ and $L_s(\alpha)$ to generate the input. Their motivation was to suppress the background pixels so that the conditioner can focus on the patterns within the object (like eyes or nose patterns within a cat class). This does not hold for our scans due to the limited texture patterns within an organ class. For example, voxel intensities within the liver are quite homogeneous with limited edges.
Thus, we feed both parts of the support set to the network and let it learn the optimal fusion that provides the best possible segmentation of the query image.


The conditioner arm has an encoder-decoder based architecture consisting of four encoder blocks, four decoder blocks, separated by a bottleneck layer, see fig.~\ref{fig:architecture}. Both encoder and decoder blocks consist of a generic block constituting a convolutional layer with kernel size of $5\times5$, stride of $1$ and $16$ output feature maps, followed by a parametric ReLU activation function~\citep{he2015delving} and a batch normalization layer. In the encoder block, the generic block is followed by a max-pooling layer of $2\times2$ and stride 2, which reduces the spatial dimension by half. In the decoder block, the generic block is preceded by an unpooling layer~\citep{noh2015learning}. The pooling indices during the max-pool operations are stored and used in the corresponding unpooling stage of the decoder block for up-sampling the feature map. 
Not only is the unpooling operation parameter free, which reduces the model complexity, but it also aids to preserve the spatial consistency for fine-grained segmentation. 
Furthermore, it must be noted that \emph{no skip connections} are used between the encoder and decoder blocks unlike the standard U-net architecture~\citep{ronneberger2015u}. The reason for this important design choice is discussed in Sec.~\ref{sec:skip_conn}.


\subsubsection{Interaction Block using `Squeeze \& Excitation' modules}
The interaction blocks play a key role in the few-shot segmentation framework. These blocks take the \emph{task representation} of the conditioner as input and convey them to the segmenter to steer segmentation of the query image. Ideally these blocks should: (i) be light-weight to only marginally increase the model complexity and computation time, and (ii) ease training of the network by improving gradient flow. 

We use the recently introduced `Squeeze \& Excitation' (SE) modules for this purpose. SE modules are computational units to achieve adaptive re-calibration of  feature maps within any CNN~\citep{hu2018squeeze}.  SE blocks can boost the performance of CNNs, while increasing model complexity only marginally. 
For classification~\citep{hu2018squeeze}, the feature maps are spatially squeezed to learn a channel descriptor, which is used to excite (or re-calibrate) the feature map, emphasizing certain important channels. We refer to it as spatial squeeze and channel excitation block (cSE). 
In our recent work, we extended the idea to segmentation, where re-calibration was performed by squeezing channel-wise and exciting spatially (sSE), emphasizing relevant spatial locations~\citep{roy2018recalibrating, roy2018concurrent}. 
In both cases, SE blocks are used for self re-calibration, i.e, the same feature map is used as input for squeezing and excitation operations. However, here we propose to use SE blocks for the interaction between the conditioner and the segmenter. The conditioner feature maps are taken as input for the squeezing operation and its outputs are used to excite the segmentation feature maps as detailed below.

\paragraph{Channel Squeeze \& Spatial Excitation (sSE)}
The sSE block squeezes a conditioner feature map $\mathbf{U}_{\mathrm{con}} \in \mathbb{R}^{H \times W \times C^\prime}$ along the channels and excites the corresponding segmenter feature map $\mathbf{U}_{\mathrm{seg}} \in \mathbb{R}^{H \times W \times C}$ spatially, conveying the information from the support set to aid the segmentation of the query image. 
$H$, $W$ are the height and width of feature maps, $C^\prime$ and $C$ are the number of channels for the conditioner and the segmenter feature maps, respectively. 
Here, we consider a particular slicing strategy to represent the input tensor $\mathbf{U}_{\mathrm{con}} = [\mathbf{u}^{1,1}_{\mathrm{con}}, \mathbf{u}^{1,2}_{\mathrm{con}} \dots, \mathbf{u}^{j,\iota}_{\mathrm{con}}, \dots, \mathbf{u}^{H, W}_{\mathrm{con}}]$, where $\mathbf{u}^{j,\iota}_{\mathrm{con}} \in \mathbb{R}^{1 \times 1 \times C^\prime}$ with $j \in \{ 1, 2, \dots, H \}$ and $\iota \in \{ 1, 2, \dots, W \}$.
Similarly for segmenter feature map $\mathbf{U}_{\mathrm{seg}} = [\mathbf{u}^{1,1}_{\mathrm{seg}}, \mathbf{u}^{1,2}_{\mathrm{seg}} \dots, \mathbf{u}^{j,\iota}_{\mathrm{seg}}, \dots, \mathbf{u}^{H, W}_{\mathrm{seg}}]$.
The spatial squeeze operation is performed using a convolution $\mathbf{q} = \mathbf{W}_{sq} \star \mathbf{U}_{\mathrm{con}}$ with $\mathbf{W}_{sq} \in \mathbb{R}^{1 \times 1 \times C^\prime}$, generating a projection tensor $\mathbf{q} \in \mathbb{R}^{H \times W}$. 
This projection $\mathbf{q}$ is passed through a sigmoid gating layer $\sigma(\cdot)$ to rescale activations to $[0, 1]$, which is used to re-calibrate or excite $\mathbf{U}_{\mathrm{seg}}$ spatially to generate

\begin{multline}
    \hat{\mathbf{U}}_{\mathrm{seg}} = [\sigma(q_{1,1})\mathbf{u}^{1,1}_{\mathrm{seg}}, \dots, \sigma(q_{j,k})\mathbf{u}^{j,\iota}_{\mathrm{seg}}, \\  \dots, \sigma(q_{H,W})\mathbf{u}^{H, W}_{\mathrm{seg}}].
\end{multline}
\noindent
The architectural details of this module are presented in fig.~\ref{fig:squeeze}.

\subsubsection{Segmenter Arm}
The goal of the segmenter arm is to segment a given query image $I_q$ with respect to a new, unknown class $\alpha$, by using the information passed by the conditioner, which captures a high-level information about the previously unseen class $\alpha$.
The sSE modules in the interaction block compresses the \emph{task representation} of the conditioner and adaptively re-calibrate the segmenter's feature maps by spatial excitation.

The encoder-decoder architecture of the segmenter is similar to the conditioner, with a few differences. Firstly, the convolutional layers of both the encoder and decoder blocks in the segmenter have $64$ output feature maps, in contrast to $16$ in the conditioner. This provides the segmenter arm with a higher model complexity than the conditioner arm. We will justify this choice in Sec.~\ref{sec:model_complexity}. Secondly, unlike the conditioner arm, the segmenter arm provides a segmentation map as output, see Fig.~\ref{fig:architecture}. Thus a classifier block is added, consisting of a $1\times1$ convolutional layer with $2$ output feature maps (foreground, background), followed by a soft-max function for inferring the segmentation. Thirdly, in the segmenter, after every encoder, decoder and bottleneck block, the interaction block re-calibrates the feature maps, which is not the case in the conditioner arm.


\subsection{Training Strategy}
We use a similar training strategy to \citet{shaban2017one}. We simulate the one-shot segmentation task with the training dataset $\mathcal{D}_\text{Train}$ as described below. It consists of two stages (i) Select a mini-batch using the \emph{Batch Sampler} and (ii) Training the network using the selected mini-batch.

\paragraph{Batch Sampler}
To simulate the one-shot segmentation task during training, we require a specific strategy for selecting samples in a mini-batch that differs from traditional supervised training. For each iteration, we follow the steps below to generate batch samples:
\begin{enumerate}
    \item We first randomly sample a label $\alpha \in \mathcal{L}_\text{Train}$.
    \item Next, we randomly select $2$ image slices and their corresponding label maps, containing the semantic label $\alpha$, from training data $\mathcal{D}_\text{Train}$.
    \item The label maps are binarized representing semantic class $\alpha$ as foreground and the rest as background.
    \item One pair constitutes the support set $(I_s, L_s(\alpha))$ and the other pair the query set $(I_q, L_q(\alpha))$, where $L_q(\alpha)$ serves as ground truth segmentation for computing the loss.
\end{enumerate}

\paragraph{Training}
The network receives the support pair $(I_s, L_s(\alpha))$ and the query pair $(I_q, L_q(\alpha))$ as a batch for training purpose. The support pair $(I_s, L_s(\alpha))$ is concatenated and provided as 2-channeled input to the conditioner arm.
The query image $I_q$ is provided as input to the segmentation arm. 
With these inputs to the two arms, one feed-forward pass is performed to predict the segmentation $M_q(\alpha)$ for the query image $I_q$  for label $\alpha$. We use the Dice loss~\citep{milletari2016v} as the cost function, which is computed between the prediction $M_q(\alpha)$ and the ground truth $L_q(\alpha)$ as
\begin{equation}
    \mathcal{L}_{\mathrm{Dice}} = 1 - \frac{2 \sum_{\mathbf{x}} M_q(\mathbf{x}) L_q(\mathbf{x})}{\sum_{\mathbf{x}} M_q(\mathbf{x}) + \sum_{\mathbf{x}} L_q(\mathbf{x})} 
\end{equation}
\noindent
where $\mathbf{x}$ corresponds to the pixels of the prediction map. 
The learnable weight parameters of the network are optimized using stochastic gradient descent (SGD) with momentum to minimize $\mathcal{L}_{\mathrm{Dice}}$. At every iteration, the batch sampler provides different samples corresponding to different $\alpha$ and the loss is computed for that specific $\alpha$ and weights are updated accordingly. With the target class $\alpha$ keeps changing at every iteration, the network converges. Thus, after convergence we can say that the prediction becomes agnostic to the chosen $\alpha$. That is for a new $\alpha$, the network should be able to perform segmentation, which is what we expect during inference of a one-shot segmentation framework.

\begin{figure*}[h]
\center{\includegraphics[width=0.85\textwidth]{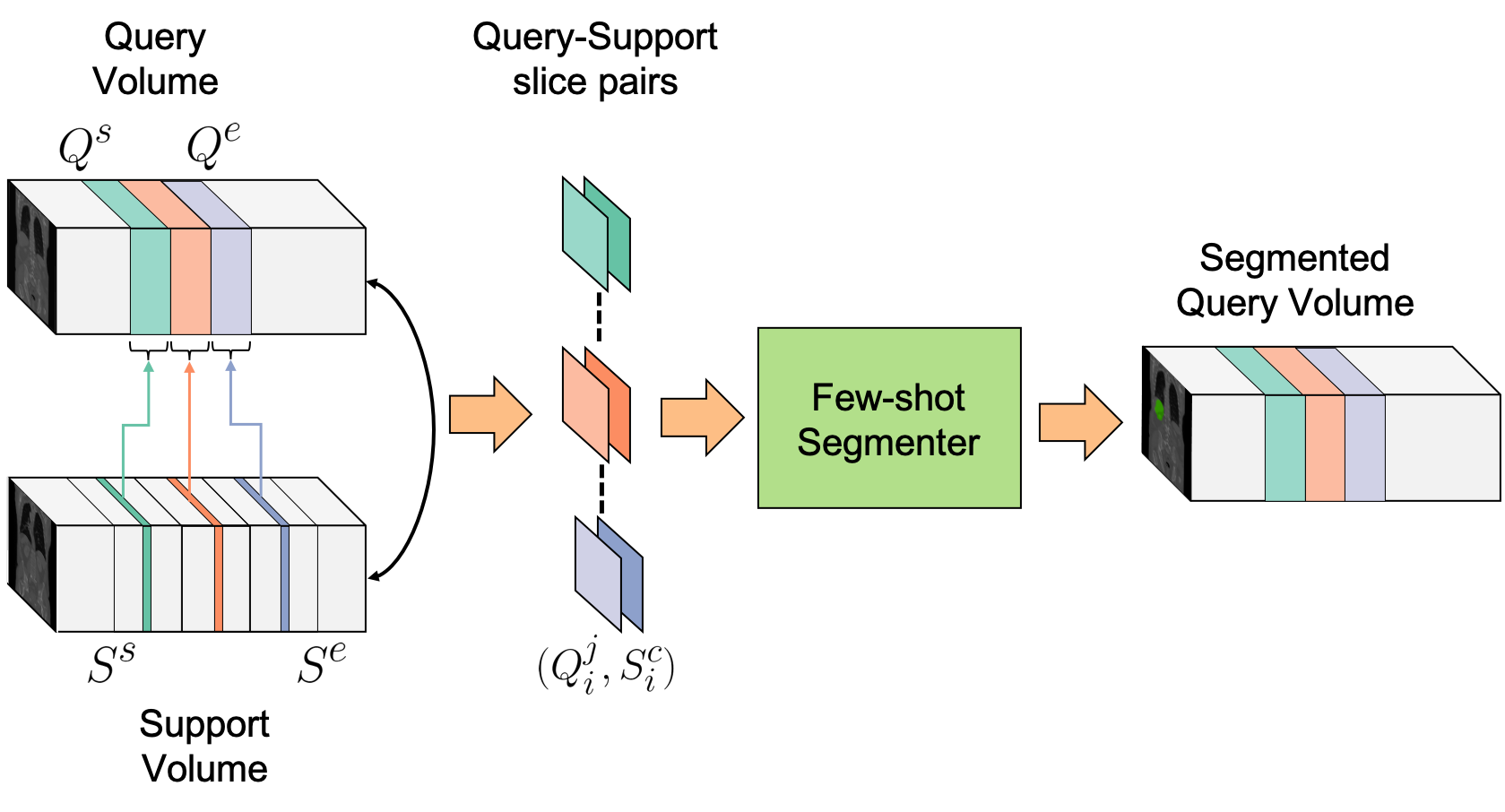}}
\caption{Illustration of the few-shot volumetric segmentation strategy for $k=3$.
We divide both the query volume and support volume into $k$ group of slices. The annotated center slice of the $i^{th}$ group in the support volume is paired with all the slices of $i^{th}$ group of query volume to infer their segmentation. This is done for $i \in \{1,2,3\}$ and is passed to the few-shot segmenter for segmenting the whole volume.}
\label{fig:eval}
\end{figure*}

\subsection{Volumetric Segmentation Strategy}
As mentioned in the previous section, the network is trained with 2D images as support set and query.
But, during the testing phase, a 3D query volume needs to be segmented.
Therefore, from the support volume, we need to select a sparse set of annotated slices that form the support set.
A straightforward extension for segmenting the query volume is challenging as there is no established strategy to pair the above selected support slices to all of slices of the query volume, which would yield the best possible segmentation. In this section, we propose a strategy to tackle this problem.

Assume we have a budget of annotating only $k$ slices in the support volume, a query volume is segmented following procedure:
\begin{enumerate}
    \item Given a semantic class, we first indicate the range of slices (along a fixed orientation) where the organ lies for both support and query volume. Let us assume the ranges are $[S^s, S^e]$ for the support and $[Q^s, Q^e]$ for the query volume. 
    Here the superscript indicates the start $s$ and end $e$ slice indices.
    \item Based on the budget $k$, both ranges $[S^s, S^e]$ and $[Q^s, Q^e]$ are divided into $k$ equi-spaced groups of slices. Let us indicate the groups by $[\{S_1^i\}, \dots, \{S_k^i\}]$ and $[\{Q_1^i\}, \dots, \{Q_k^i\}]$ respectively. Here the subscript indicates the group number.
    \item In each of the $k$ support volume groups, center slices  $[S_1^c, \dots, S_k^c]$ are annotated to serve as the support set. 
    \item We pair the annotated center slice $S_j^c$ with all the slices of the group $\{Q_j^i\}$ for all $i \in \{ 1, \dots, k \}$. This forms the input for the segmenter and the conditioner to generate the final volume segmentation.
\end{enumerate}

The overall process of volumetric evaluation is illustrated in Fig.~\ref{fig:eval}.
In our experiments, we observed that if the support slice and query slice are similar, segmentation performance is better than if they were very dissimilar. 
Therefore, it is beneficial if the overall contrast of the scans (i.e. the intensity values or mutual information) is similar.
This can be intuitively understood as the quality of the support the slice has a major impact on the segmenter's performance. In our evaluation strategy, for a fixed budget $k$, we made sure that the dissimilarity between the support slice and the corresponding query slice is minimal. 
It must be noted that in the evaluation strategy $[Q^s, Q^e]$ must be provided for the query volume. In our experiments, we pre-computed them using the label mask of the target organ. In practice, this could be done either manually by quickly scanning the slices, or using a simple automated tool that can be trained for this purpose.

\section{Dataset and Experimental Setup}
\label{sec:exp_setup}
\subsection{Dataset Description}
We choose the challenging task of organ segmentation from contrast-enhanced CT (ceCT) scans, for evaluating our few-shot volumetric segmentation framework. We use the Visceral dataset~\citep{jimenez2016cloud}, which consists of two parts (i) silver corpus (with 65 scans) and (ii) gold corpus (20 scans). All the scans were resampled to a voxel resolution of $2\mathrm{mm}^3$.

\subsection{Problem Formulation}
As there is no existing benchmark for few-shot image segmentation on volumetric medical images, we formulate our own experimental setup for the evaluation. We use the silver corpus scans for training ($\mathcal{D}_\text{Train}$). For testing, we use the gold corpus dataset. One volume is used to create the support set (Volume ID: \texttt{10000132\_1\_CTce\_ThAb}), $14$ volumes were used as validation set and $5$ volumes as test set. The IDs of the respective volumes are reported at the end of the manuscript. In the experiments presented in Sec.~\ref{sec:se_inter} to Sec.~\ref{sec:num_support}, we use the validation set as we use these results to determine the architectural configuration, and number of support slices. Finally, we use these results and compare against existing approaches on the test set in Sec.~\ref{sec:compare}.

We consider the following six organs as semantic classes in our experiments: 
\begin{enumerate}
    \item Liver
    \item Spleen
    \item Right Kidney
    \item Left Kidney
    \item Right Psoas Muscle
    \item Left Psoas Muscle
\end{enumerate}

We perform experiments with 4 Folds, such that each organ is considered as an unknown semantic class once per-fold. The training and testing labels for each of the folds are reported in Tab.~\ref{tab:folds}.

\begin{table}[]
\centering
\caption{Semantic labels used for training and testing in all the experimental folds. Left and Right are abbreviated as L. and R. Psoas Muscle is abbreviated as P.M.}
\rowcolors{2}{gray!20}{white}
\begin{tabular}{lllll}
 \toprule
 & Fold 1 & Fold 2 & Fold 3 & Fold 4 \\
 \otoprule
 Liver        & \emph{Test}  & Train & Train & Train \\
 Spleen       & Train &  \emph{Test}  & Train & Train \\
 L./R. Kidney & Train & Train &  \emph{Test}  & Train \\
 L./R. P. M.  & Train & Train & Train &  \emph{Test} \\
 \bottomrule
\end{tabular}
\label{tab:folds}
\end{table}

\subsection{Hyperparameters for Training the Network}
Due to the lack of pre-trained models, we could not use the setup from \citet{shaban2017one} for training. 
Thus, we needed to define our own hyperparameter settings, listed in Table~\ref{tab:hyperparams}. Please note that the hyperparameters were estimated by manually trying out different combinations, rather than employing a hyperparameter optimization framework, which could lead to better results but at the same time is time-consuming.

\begin{table}[]
\centering
\caption{List of hyperparameters used for training the few-shot segmenter.}
\rowcolors{2}{gray!20}{white}
\begin{tabular}{lr}
\toprule
Hyperparameter & Value \\
\otoprule
 Learning Rate & $0.01$\\
 Weight decay constant & $10^{-4}$ \\
 Momentum & $0.99$  \\
 No. of epochs & $10$ \\
 Iterations per epoch & $500$ \\
\bottomrule
\end{tabular}
\label{tab:hyperparams}
\end{table}



\section{Experimental Results and Discussion}
\label{sec:results}

\subsection{`Squeeze \& Excitation' based Interaction}
\label{sec:se_inter}
In this section, we investigate the optimal positions of the SE blocks for facilitating interaction and compare the performance of cSE and sSE blocks. Here, we set the number of convolution kernels of the conditioner arm to $16$ and the segmenter arm to $64$. We use $k=12$ support slices from the support volume. Since the aim of this experiment is to evaluate the position and the type of SE blocks, we keep the above parameters fixed, but evaluate them later.
With four different possibilities of placing the SE blocks and two types cSE or sSE, we have a total of 8 different baseline configurations.
The configuration of each of these baselines and their corresponding segmentation performance per fold is reported in Tab.~\ref{tab:pos_se_bl}. 

Firstly, one observes that BL-1, 3, 5, 7 with sSE have a decent performance (more than $0.4$ Dice score), whereas BL-2, 4, 6, 8 have a very poor performance (less than $0.1$ Dice score). This demonstrates that sSE interaction modules are far superior to cSE modules in this application of few-shot segmentation. 
It is very difficult to understand the dynamics of the network to say for certain why such a behavior is observed. Our intuition is that the under-performance using channel SE blocks is associated with the global average pooling layer it uses, which averages the spatial response to a scalar value. In our application (or medical scans in general), the target semantic class covers a small proportion of the support slice (around 5-10\%). When averaged over all the pixels, the final value is highly influenced by the background activations. The role of the interaction blocks is to convey the target class's semantic information from conditioner to segmenter. By using channel SE as global average pooling the class information is mostly lost, thus cannot convey the relevant information to the segmenter.

The second conclusion from Tab.~\ref{tab:pos_se_bl} is that out of all the possible positions of the interaction block, BL-7, i.e., sSE blocks between all encoder, bottleneck and decoder blocks, achieved the highest Dice score of $0.567$. This result is consistent across all folds. 
BL-7 outperformed the remaining baselines for Fold-1 (liver), Fold-2 (spleen), Fold-3 (L/R kidney) and Fold-4 (L/R psoas muscle) by a margin of 0.1 to 0.8 Dice points. This might be related to the relative difficulty associated with each organ. Due to the contrast and size, the liver is relatively easy to segment in comparison to spleen, kidney, and psoas muscles. 
Also, BL-1, 3 and 5 performed poorly in comparison to BL-7. This indicates that more interactions aids in better training. Comparing BL-1, BL-3 and Bl-5, we observe that BL-1 provides better performance. This indicates that encoder interactions are much powerful than bottleneck or decoder interactions. But, as BL-7 has a much higher performance than BL-1, BL-3 and BL-5, we believe that encoder, bottleneck and decoder interactions provide complementary information to the segmenter for more accurate query segmentation.
From these results, we conclude that interaction blocks based on sSE are most effective and we use sSE-based interactions between all encoder, bottleneck, and decoder blocks
in subsequent experiments.

\begin{table*}[t]
\centering
\caption{The performance of our few-shot segmenter (per-fold and mean Dice score) by using either sSE or cSE module, at different locations (encoder, bottleneck and decoder) of the network. Left and Right are abbreviated as L. and R. Psoas Muscle is abbreviated as P.M.}
\rowcolors{3}{gray!20}{white}
\begin{tabular}{lcccccrrrrr}
\toprule
\hiderowcolors
& \multicolumn{3}{c}{Position of SE} & \multicolumn{2}{c}{Type of SE} & \multicolumn{5}{c}{Dice Score on Validation set} \\ \cmidrule(r){2-4} \cmidrule(lr){5-6} \cmidrule(l){7-11}
 & Encoder & Bottleneck & Decoder & Spatial & Channel & Liver & Spleen & L/R kidney & L/R P.M. & Mean  \\
 \otoprule
 \showrowcolors
 BL-1 & \checkmark & $\times$ & $\times$ & \checkmark & $\times$ & $0.667$ & $0.599$ & $0.385$ & $0.339$ & $0.497$ \\
 BL-2 & \checkmark & $\times$ & $\times$ & $\times$ & \checkmark & $0.086$ & $0.032$ & $0.087$ & $0.017$ & $0.056$ \\
 BL-3 & $\times$ & \checkmark & $\times$ & \checkmark & $\times$ & $0.680$ & $0.398$ & $0.335$ & $0.252$ & $0.416$ \\
 BL-4 & $\times$ & \checkmark & $\times$ & $\times$ & \checkmark & $0.060$ & $0.018$ & $0.090$ & $0.032$ & $0.050$ \\
 BL-5 & $\times$ & $\times$ & \checkmark & \checkmark & $\times$ & $0.683$ & $0.534$ & $0.278$ & $0.159$ & $0.414$ \\
 BL-6 & $\times$ & $\times$ & \checkmark & $\times$ & \checkmark & $0.051$ & $0.014$ & $0.010$ & $0.003$ & $0.020$ \\
 BL-7 & \checkmark & \checkmark & \checkmark & \checkmark & $\times$ & $0.700$ & $0.607$ & $0.464$ & $0.499$ & $\mathbf{0.567}$ \\
 BL-8 & \checkmark & \checkmark & \checkmark & $\times$ & \checkmark & $0.026$ & $0.003$ & $0.001$ & $0.001$ & $0.008$ \\
\bottomrule 
\end{tabular}
\label{tab:pos_se_bl}
\end{table*}

\begin{figure*}[t]
\center{\includegraphics[width=\textwidth]{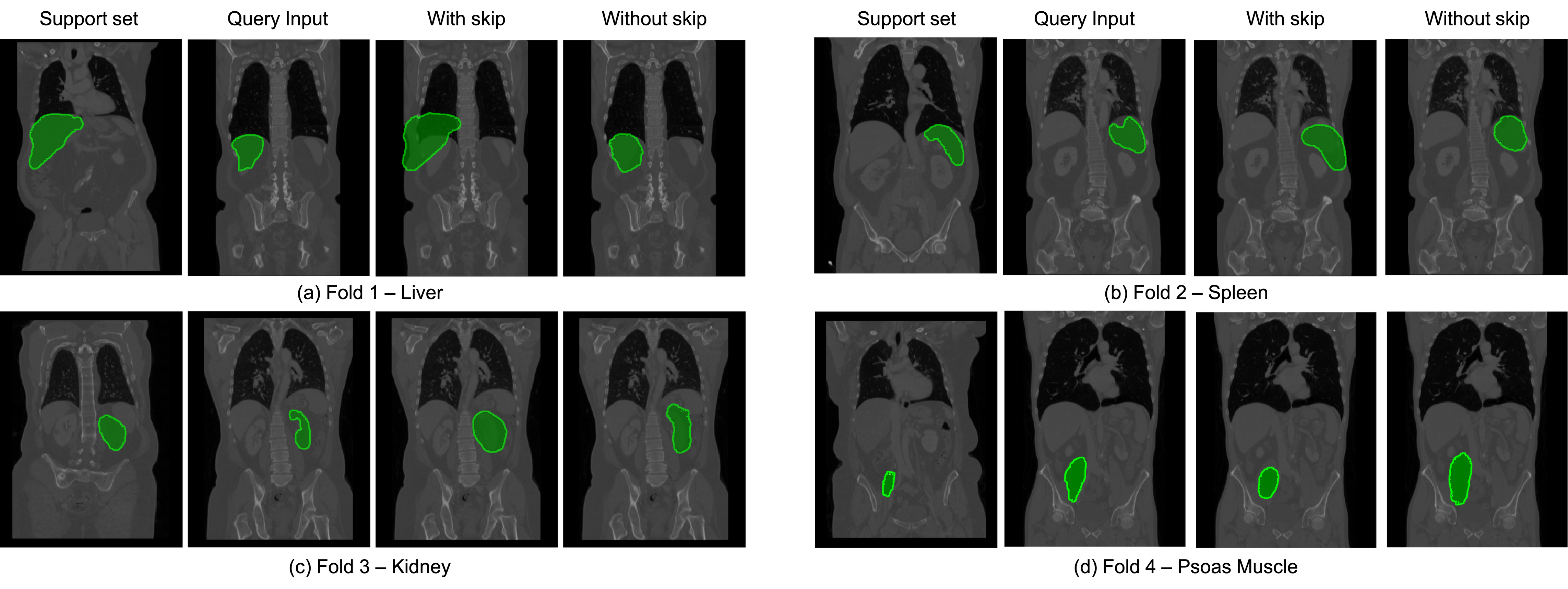}}
\caption{Qualitative results of few-shot segmenter with and without skip connections to demonstrate the \emph{copy over effect}. The sub-figures (a-d) refer to the examples from each of the folds namely liver, spleen, left kidney and right psoas muscles, respectively. For each sub-figure, the first column indicates the support image with the manual outline of the organ, the second column indicates the query image with manual annotation, the third column indicates the prediction of the query image with skip connection, and the fourth column indicates the prediction of the query image without skip connections (proposed approach). All annotations are shown in green. A clear \emph{copy over effect} can be observed for all the folds when analyzing the mask of the support annotation and the prediction with skip connections.}
\label{fig:copyover}
\end{figure*}

\subsection{Effect of Skip Connections in the Architecture}
\label{sec:skip_conn}
Due to the success of the U-net architecture~\citep{ronneberger2015u}, using skip connections in F-CNN models has become a very common design choice.
With skip connections, the output feature map of an encoder block is concatenated with the input of the decoder block with an identical spatial resolution. In general, this connectivity aids in achieving a superior segmentation performance as it provides a high contextual information in the decoding stage and facilitates gradient flow. In our experiments, we intuitively started off with having skip connections in both the conditioner arm and the segmenter arm, but observed an unexpected behavior in the predicted query segmentation masks. By including skip connections, the network mostly copies the binary mask of the support set to the output. This is observed for all the folds both in train and test set. We refer to this phenomenon as the \emph{copy over effect}. Qualitative examples are illustrated for each fold in Fig~\ref{fig:copyover}, where we see that, despite of the support and the query images having different shapes, the prediction on the query image is almost identical to the support binary mask.
We also performed a quantitative analysis to observe the effect on Dice scores due to this \emph{copy over effect}. Table~\ref{tab:skip_conn} reports the performance with and without skip connections, where we observe  a 3\% decrease in Dice points due to the addition of skip connections. We also performed experiments by separately adding the skip connections in the conditioner and the segmenter arm. We observe that the inclusion of skip connections only in the conditioner arm reduced the performance by $6\%$ Dice points, whereas adding them only in the segmenter arm made the training unstable.
For this evaluation, the number of convolution kernels for conditioner and segmenter were fixed at $16$ and $64$, respectively, and the evaluation was conducted with $k=12$ support slices.

\begin{table*}[t]
\centering
\caption{The segmentation performance (per-fold and mean Dice score) on test scans, with and without using skip connections within our few-shot segmenter. Left and Right are abbreviated as L. and R. Psoas Muscle is abbreviated as P.M.}
\rowcolors{2}{gray!20}{white}
\begin{tabular}{ccrrrrr}
\toprule
 \multicolumn{2}{c}{Skip Connections} & \multicolumn{5}{c}{Dice Score on Validation set} \\
 \cmidrule(r){1-2} \cmidrule(l){3-7}
 Conditioner & Segmenter & Liver & Spleen & L/R kidney & L/R P.M. & Mean \\
 \otoprule
 $\times$ & $\times$ & $0.700$ & $0.607$ & $0.464$ & $0.499$ & $\mathbf{0.567}$ \\
 \checkmark & $\times$ & $0.561$ & $0.495$ & $0.457$ & $0.447$ & $0.505$ \\
 $\times$ & \checkmark & $0.096$ & $0.026$ & $0.025$ & $0.019$ & $0.042$ \\
 \checkmark & \checkmark & $0.561$ & $0.549$ & $0.543$ & $0.501$ & $0.538$ \\
 \bottomrule
\end{tabular}
\label{tab:skip_conn}
\end{table*}

\subsection{Model Complexity of the Conditioner Arm}
\label{sec:model_complexity}
One important design choice is to decide the relative model complexity of the conditioner arm compared to the segmenter arm. As mentioned in Sec.~\ref{sec:backgnd}, the conditioner takes in the support example and learns to generate \emph{task representations}, which are passed to the segmenter arm through interaction blocks. This is utilized by the segmenter to segment the query image. 
We fix the number of kernels of the convolutional layers (for every encoder, bottleneck, and decoder) for the segmenter arm to $64$.
We use this setting as this has proven to work good in our prior segmentation works across different datasets~\citep{roy2019quicknat, roy2018recalibrating}.
Next, we vary the number of kernels of the conditioner arm to $\{8, 16, 32, 64 \}$. The number of support slices remains fixed to $k=12$. We report the segmentation results of these settings in Table~\ref{tab:cond_model_comp}. The best performance was observed for the conditioner with $16$ convolution kernels. 
One possible explanation of this could be that too low conditioner complexity (like $8$) leads to a very weak \emph{task representation}, thereby failing to reliably supporting to the segmenter arm. Whereas higher conditioner arm complexity, $32$ and $64$ kernels (same as segmenter complexity), might lead to improper training due to increased complexity under limited training data and interaction. 
We fix the number of conditioner convolution kernels to $16$ in our following experiments. 


\begin{table*}[t]
\centering
\caption{Effect of model complexity of the conditioner arm (Number of convolution kernels) on segmentation performance, provided a fixed model complexity (Number of convolution kernels fixed to $64$) of the segmenter arm. Left and Right are abbreviated as L. and R. Psoas Muscle is abbreviated as P.M.}
\rowcolors{2}{gray!20}{white}
\begin{tabular}{crrrrr}
\otoprule
\hiderowcolors
 \multicolumn{1}{c}{Channels in} & \multicolumn{5}{c}{Dice Score on Validation set} \\ \cmidrule{2-6}
Conditioner Arm & Liver & Spleen & L/R kidney & L/R P.M. & Mean \\
 \otoprule
  \showrowcolors
 $8$ & $0.628$ & $0.275$ & $0.429$ & $0.276$ & $0.402$ \\
 $16$ & $0.700$ & $0.607$ & $0.464$ & $0.499$ & $\mathbf{0.567}$ \\
 $32$ & $0.621$ & $0.551$ & $0.378$ & $0.280$ & $0.457$ \\
 $64$ & $0.659$ & $0.417$ & $0.421$ & $0.247$ & $0.436$ \\
 \bottomrule
\end{tabular}
\label{tab:cond_model_comp}
\end{table*}

\subsection{Effect of the number of Support Slice Budget}
\label{sec:num_support}
In this section, we investigate the performance when changing the budget for the number of support slices $k$ selected from the support volume for segmenting all the query volumes. Here, $k$ can be thought of as the `number of shots' for volumetric segmentation. In all the previous experiments we fix $k=12$. Here, we vary $k$ between $\{1, 3, 5, 7, 10, 12, 15, 17, 20\}$ and report the per-fold and overall mean segmentation performance on validation set in Table~\ref{tab:effect_support_slice}. The per-fold performance analysis reveals that the minimum number of slices needed for a decent accuracy varies with the size of the target organ to be segmented. 

For Fold-1 (liver), one-shot volume segmentation ($k=1$) yielded a Dice score of $0.678$, which increased to $0.701$ with $k=20$. 
We observed a saturation in performance (Dice score of $0.70$) with only $12$ slices. 
The segmentation performance only marginally increased with higher values of $k$.  
For Fold-2 (spleen), the segmentation performance initially increases with the increase in the value of $k$, then the performance saturates with $k \ge 10$ at a Dice score of $0.60$. The spleen is more difficult to segment than liver, thus requires more support.
For Fold-3 (right/ left kidney), we observe behavior similar to Fold-2. The segmentation performance increases initially with increase in the value of $k$ and then saturates at a Dice score of $0.46$ (this is the mean between the two classes, left and the right kidney) at $k\ge10$.
Also for Fold-4 (right/ left psoas muscle), we see the Dice score saturates at $0.50$ for $k=10$. 
The overall mean Dice score across all the folds also saturates at $0.56$ with $k=10$.

Based on these results, we conclude that $k=10$ is the maximum number of support slices required for our application. Thus, we use this configuration in the next experiments.

We also report in Tab.~\ref{tab:slice_range} the mean number of slices in the testing volumes for each organ to indicate of how sparse the slices were selected for volumetric evaluation.

\begin{table}[h]
\centering
\caption{Extent of slices (for coronal axis) for different target organs in the Visceral dataset.}
\rowcolors{2}{gray!20}{white}
\begin{tabular}{lc}
\toprule
 Organs & Extent of Slices \\
 \otoprule
 Liver & $106 \pm 12$ \\
 Spleen & $50 \pm 8$ \\
 R. Kidney & $34 \pm 4$ \\
 L. Kidney & $36 \pm 5$ \\
 R. P.M. & $31 \pm 5$ \\
 L. P.M & $31 \pm 3$ \\
 \bottomrule
\end{tabular}
\label{tab:slice_range}
\end{table}

\begin{table*}[t]
\centering
\caption{The segmentation performance (per-fold and mean Dice score) on validation scans, by varying the number of annotated slice ($k$) as support in the support volume. Left and Right are abbreviated as L. and R. Psoas Muscle is abbreviated as P.M.}
\rowcolors{3}{gray!20}{white}
\begin{tabular}{crrrrr}
\toprule
\hiderowcolors
 No. of support & \multicolumn{5}{c}{Dice Score on Validation set} \\ \cmidrule{2-6}
 \multicolumn{1}{c}{slices ($k$)} & Liver & Spleen & L/R kidney & L/R P.M. & Mean \\
\otoprule
\showrowcolors
  $1$ & $0.678$ & $0.503$ & $0.385$ & $0.398$ & $0.491$ \\
 $3$ & $0.692$ & $0.490$ & $0.422$ & $0.437$ & $0.510$ \\
 $5$ & $0.685$ & $0.557$ & $0.445$ & $0.496$ & $0.546$ \\
 $7$ & $0.694$ & $0.577$ & $0.457$ & $0.507$ & $0.559$ \\
 $10$ & $0.688$ & $0.600$ & $0.466$ & $0.505$ & $0.565$ \\
 $12$ & $0.700$ & $0.607$ & $0.464$ & $0.499$ & $0.567$ \\
 $15$ & $0.700$ & $0.607$ & $0.464$ & $0.496$ & $0.567$ \\
 $17$ & $0.700$ & $0.609$ & $0.465$ & $0.497$ & $0.567$ \\
 $20$ & $0.701$ & $0.606$ & $0.468$ & $0.496$ & $0.568$ \\
 \bottomrule
\end{tabular}
\label{tab:effect_support_slice}
\end{table*}

\begin{table*}[h]
\centering
\caption{Comparison of our proposed few-shot segmenter against the existing methods. For each method, per-fold and mean Dice score and average surface distance (in mm) are reported for the test set. Left and Right are abbreviated as L. and R. Psoas Muscle is abbreviated as P.M. $^\star$Classifier Regression~\citep{shaban2017one} training resulted in mode-collapse, hence no Dice score is reported. Feature Fusion is abbreviated to F.F. and Classifier Regression to C.R.}
\rowcolors{3}{gray!20}{white}
\begin{tabular}{lrrrrr}
\toprule
\hiderowcolors
 & \multicolumn{5}{c}{Dice Score on Test set} \\ \cmidrule{2-6}
 Method & Liver & Spleen & L/R kidney & L/R P.M. & Mean \\
 \otoprule
 \showrowcolors
 Proposed & $0.680$ & $0.475$ & $0.338$ & $0.450$ & $\mathbf{0.485}$ \\
 C.R.$^\star$~(Adapted from \cite{shaban2017one}) & $-$ & $-$ & $-$ & $-$ & $-$ \\
 F.F.~(Adapted from \cite{rakelly2018few}) & $0.247$ & $0.267$ & $0.307$ & $0.258$ & $0.270$ \\
 F.F. + C.R. & $0.224$ & $0.197$ & $0.348$ & $0.411$ & $0.295$ \\ 
 Fine-Tuning~\citep{caelles2017one} & $0.307$ & $0.016$ & $0.003$ & $0.043$ & $0.092$ \\
 \bottomrule
 \hiderowcolors
 & \multicolumn{5}{c}{Average Surface Distance on Test set in mm} \\ \cmidrule{2-6}
 Method & Liver & Spleen & L/R kidney & L/R P.M. & Mean \\
 \otoprule
 \showrowcolors
 Proposed & $14.98$ & $10.71$ & $7.12$ & $9.13$ & $\mathbf{10.48}$ \\
 C.R.$^\star$~(Adapted from \cite{shaban2017one}) & $-$ & $-$ & $-$ & $-$ & $-$ \\
 F.F.~(Adapted from \cite{rakelly2018few}) & $32.25$ & $18.24$ & $17.16$ & $12.35$ & $20.00$ \\
 F.F. + C.R. & $38.71$ & $17.60$ & $12.64$ & $10.60$ & $19.88$ \\ 
 Fine-Tuning~\citep{caelles2017one} & $26.35$ & $-$ & $-$ & $-$ & $-$ \\
 \bottomrule
\end{tabular}
\label{tab:comp_methods}
\end{table*}

\subsection{Comparison with existing approaches}
\label{sec:compare}
In this section, we compare our proposed framework against the other existing few-shot segmentation approaches. It must be noted that all of the existing methods were proposed for computer vision applications and thus cannot directly be compared against our approach as explained in Sec.~\ref{sec:challenges}. Hence, we modified each of the existing approaches to suit our application. The results are summarized in Table~\ref{tab:comp_methods}. Also, we evaluate the results on the $5$ test query volumes.

First, we try to compare against~\citet{shaban2017one}. 
Their  main contribution was that the conditioner arm  regresses the convolutional weights, which are used by the classifier block of the segmenter to infer the segmentation of the query image. As we do not have any pre-trained models for our application, unlike~\cite{shaban2017one}, we use the same architecture as our proposed method for the segmenter and conditioner arms. No intermediate interactions were used other than the final classifier weight regression. We attempted to train the network on our dataset with a wide range of hyperparameters, but all the settings led to instability while training. It must be noted that one possible source of instability might be that we do not use a pre-trained model, unlike the original method. Thus, we were not able to compare our proposed method with this approach.

Next, we compare our approach to~\citet{rakelly2018few}. 
Again, this approach is  not directly comparable to our approach due to the lack of a pre-trained model. One of the main contributions of their approach was the interaction strategy between the segmenter and the conditioner using a technique called \emph{feature fusion}. They tiled the feature maps of the conditioner and concatenated them with the segmenter feature maps. Their implementation introduced the interaction only at a single location (bottleneck). We tried the same configuration, but the network did not converge. Thus, we modified the model by introducing the concatenation based feature fusion (instead of our sSE modules) at multiple locations between the conditioner and segmenter arms. As we have a symmetric architecture no tiling was needed. Similar to our proposed approach, we introduced this feature fusion based interaction at every encoder, bottleneck, and decoder block. In this experiment, we are comparing our spatial SE based interaction approach to the concatenation based feature fusion approach. The results are reported in Table~\ref{tab:comp_methods}. 
We observe $21\%$ higher Dice points and $10$\,mm lower average surface distance for our approach.

Next, we attempted to create hybrid baselines by combining the above adapted feature fusion approach~\citep{rakelly2018few} with classifier weight regression approach~\citep{shaban2017one}. We observe that by doing so the performance increased by $3\%$ Dice points. Still, it had a much lower Dice score ($18\%$ Dice points) in comparison to our proposed approach.

As a final baseline, we compare our proposed framework against the fine-tuning strategy similar to~\cite{caelles2017one}. For a fair comparison, we only use the silver corpus scans $(\mathcal{D}_{\mathrm{Train}})$ and $10$ annotated slices from the support volume (\texttt{10000132\_1\_CTce\_ThAb}) for training. As an architectural choice, we use our segmenter arm without the SE blocks. We pre-train the model using $\mathcal{D}_{\mathrm{Train}}$ to segment the classes of $\mathcal{L}_{\mathrm{Train}}$. After pre-training, we use the learnt weights of this model for initialization of all the layers, except for the classifier block. Then, we fine-tune it using the $10$ annotated slices of the support volume having a new class from $\mathcal{L}_{\mathrm{Test}}$. We present the segmentation performance in Table~\ref{tab:comp_methods}. Fine-tuning was carefully performed with a low learning rate of $10^{-3}$ for $10$ epochs. The $10$ selected slices were augmented during the training process using translation (left, right, top, bottom) and rotation (-15, +15 degrees). Except for fold-1 (liver, Dice score $0.30$) all the other folds had a Dice score $<0.01$. Overall, this experiment substantiated the fact that fine-tuning under such a low-data regime is ineffective, whereas our few-shot segmemtation technique is much more effective.


\subsection{Comparison with upper bound model}
In this section, we investigate the performance of our few-shot segmentation framework to the fully supervised upper bound model. For training this upper bound model, we use all the scans of the Silver Corpus (with annotations of all target organs) and deployed the trained model on the Gold Corpus. 
We use the standard U-Net~\citep{ronneberger2015u} architecture for segmentation.  
Segmentation results are shown in Table~\ref{tab:upper_bound}.

We observe that this upper bound model has 20-40\% higher Dice points and 1-7 mm lower average surface distance in comparison to our few-shot segmentation framework. 
It must be noted that this kind of difference in performance can be expected as all slices from 65 fully annotated scans were used for training. In contrast, only 10 annotated slices from a single volume were used in our approach. If access to many fully annotated volumes is provided, it is always recommended to use fully supervised training. Whenever a new class needs to be learnt from only a few slices, our framework of few-shot segmentation framework excels.
It is also worth mentioning that this drop in performance can also be observed in the PASCAL VOC benchmark from computer vision, where the fully supervised upper bound has an IoU of 0.89 using the DeepLabv3 architecture, whereas few-shot segmentation has an IoU of 0.4 ~\citep{shaban2017one}.

\begin{table}[h]
\centering
\caption{Performance of upper bound model on the Test Set.}
\rowcolors{2}{gray!20}{white}
\begin{tabular}{lcp{2.5cm}}
\toprule
 Organ & Mean Dice score & Avg. Surface Distance (mm) \\
 \otoprule
 Liver & $0.900$ & $13.15$ \\
 Spleen & $0.824$ & $3.27$ \\
 R. Kidney & $0.845$ & $3.45$ \\
 L. Kidney & $0.868$ & $3.03$ \\
 R. P.M. & $0.685$ & $8.31$ \\
 L. P.M & $0.680$ & $7.19$ \\
 \bottomrule
\end{tabular}
\label{tab:upper_bound}
\end{table}

\subsection{Qualitative Results}
\begin{figure*}
\center{\includegraphics[width=\textwidth]{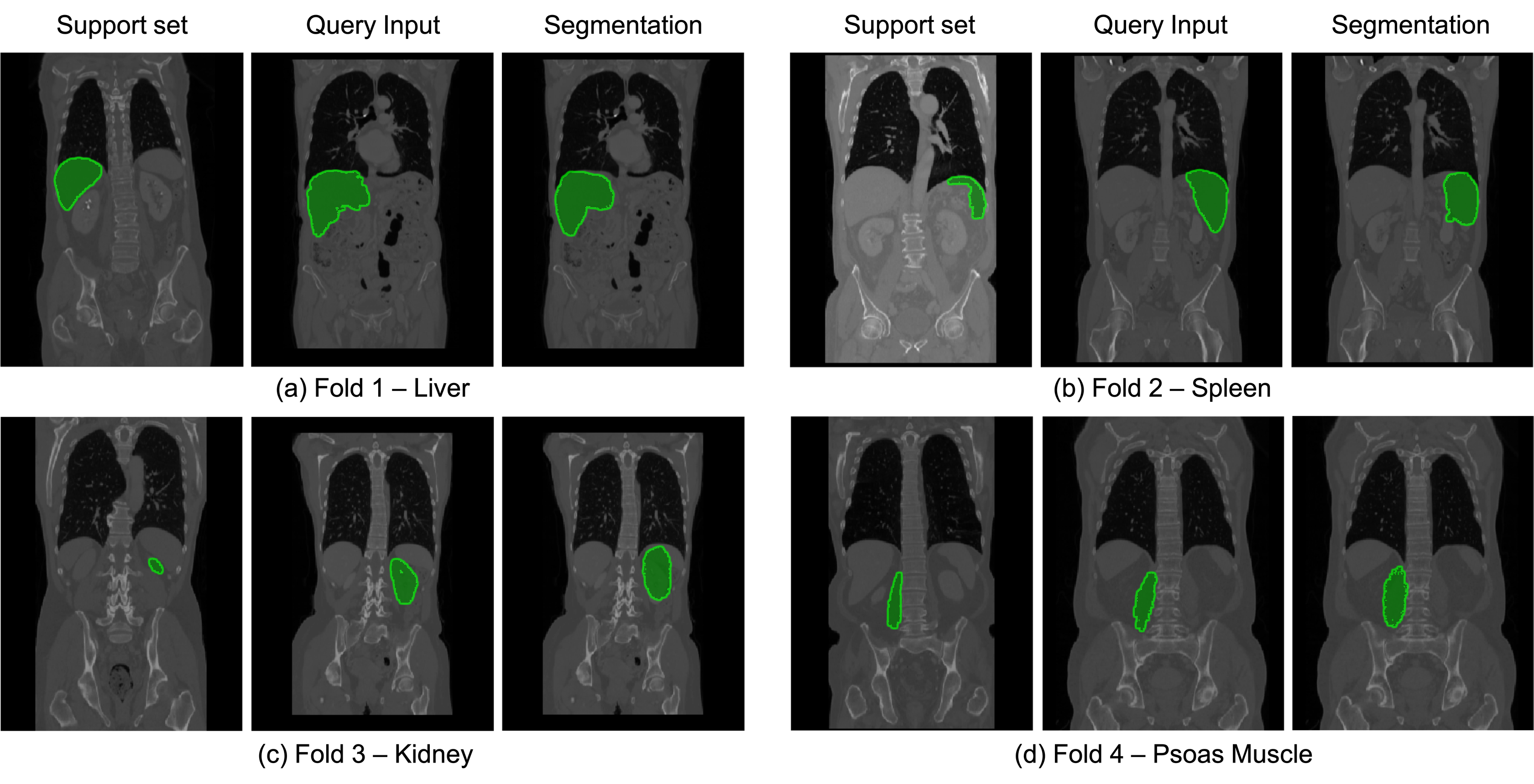}}
\caption{Qualitative results of our few-shot segmenter. The sub-figures (a-d) refer to examples from each of folds with liver, spleen, left kidney and right psoas muscles, respectively. For each of sub-figure, the first column indicates the support image with the manual outline of the organ, the second column indicates the query image with manual annotation, and the third column indicates the predicted segmentation for the query image. All the annotations are shown in green.}
\label{fig:results}
\end{figure*}

We present a set of qualitative segmentation results in Fig.~\ref{fig:results}(a-d) for folds 1-4, respectively. 
In Fig.~\ref{fig:results}(a), we show the segmentation of liver. From left to right, we present the support set with manual annotation, query input with its manual annotation, and prediction of the query input.
We observe an acceptable segmentation despite the differences in the shape and size of the liver in the support and the query slices. 
Note that the only information the network has about the organ is from a single support slice. 
In Fig.~\ref{fig:results}(b), we show a similar result for spleen. This is a challenging case where the shape of the spleen is very different in the support and query slices. Also, there is a difference in image contrast between the support and query slices. There is a slight undersegmentation of the spleen, but, considering the weak support, the segmentation is surprisingly good.
In Fig.~\ref{fig:results}(c), we present the results of left kidney. Here, we again observe a huge difference in the size of the kidney in support and query slices. The kidney appears as a small dot in the support, making it a very difficult case. 
In Fig.~\ref{fig:results}(d), we show the segmentation for right psoas muscle. In this case, the support and query slices are pretty similar to each other visually. The prediction from our framework shows a bit of over-inclusion in the psoas muscle boundary, but a decent localization and shape. 
Overall, the qualitative results visually present the effectiveness of our framework both under simple and very challenging conditions.

\begin{table*}[h]
\centering
\caption{The segmentation performance (per-fold and mean Dice score) on silver corpus {(validation set and test set combined)}, by using different volumes (Volume ID indicated in the first column) as the support volume. Left and Right are abbreviated as L. and R. Psoas Muscle is abbreviated as P.M.}
\rowcolors{2}{gray!20}{white}
\begin{tabular}{lrrrrr}
\toprule
\hiderowcolors
 Support & \multicolumn{5}{c}{Dice Score on rest of 15 volumes of the Dataset} \\ \cmidrule{2-6}
 Volume ID & Liver & Spleen & L/R kidney & L/R P.M. & Mean \\
 \otoprule
  \showrowcolors
 \texttt{10000100\_1\_CTce\_ThAb} & $0.748$ & $0.550$ & $0.445$ & $0.454$ & $0.550$ \\
 \texttt{10000106\_1\_CTce\_ThAb} & $0.690$ & $0.514$ & $0.444$ & $0.464$ & $0.528$ \\
 \texttt{10000108\_1\_CTce\_ThAb} & $0.718$ & $0.560$ & $0.406$ & $0.465$ & $0.537$ \\
 \texttt{10000113\_1\_CTce\_ThAb} & $0.689$ & $0.505$ & $0.392$ & $0.453$ & $0.510$ \\
 \texttt{10000132\_1\_CTce\_ThAb} & $0.694$ & $0.533$ & $0.369$ & $0.501$ & $0.524$ \\
 \bottomrule
\end{tabular}
\label{tab:effect_support_volume}
\end{table*}

\subsection{Dependence on Support set}
In all our previous experiments, one volume (\texttt{10000132\_1\_CTce\_ThAb}) was used as a support volume and the remaining 19 as query volumes for evaluation purposes. In this section, we investigate the sensitivity of segmentation performance on the selection of the support volume. In this experiment, we randomly choose $5$ volumes as support set from the validation set. We select one at a time and evaluate on the remaining $15$ volumes (rest of the validation set and test set combined) and report the per-fold and global Dice scores in Table~\ref{tab:effect_support_volume}. 

We observe that changing the support volume does have an effect on the segmentation performance. In Fold-1 (liver), the performance varies by $6\%$ Dice points across all the $5$ selected support volume. This change is $5\%$, $8\%$ and $5\%$ Dice points for Fold-2 (spleen), Fold-3 (R/L kidney), Fold-4 (R/L psoas muscle), respectively. The overall mean Dice scores vary by $4\%$ points. We conclude that it is important to select an appropriate support volume that is representative of the whole query set. Yet, a good strategy for making the selection remains as a future work. Nevertheless, our framework shows some robustness to the selection.

\subsection{Discussion on spatial SE as interaction blocks}
One concern regarding the use of spatial SE blocks for interaction might be the spatial alignment of the target class between the support and query images. 
Although in our application, there exist some partial overlap of the target organ between the support and query slice, we believe the sSE based interaction is also capable of handling cases where there is no such overlap.
We acknowledge that similarity in spatial location does help in our application. However that is not the only factor driving the segmentation. In Table~\ref{tab:pos_se_bl}, we present experiments for a configuration denoted as BL-3. In this design, we only keep the sSE block interaction at the bottleneck between Segmenter and Conditioner. Note that the spatial resolution in bottleneck feature map is very low (size: $16\times32$ for our case). This configuration can be considered as a spatially invariant fusion. In this scenario, we also achieve a decent segmentation score. This is further boosted by adding sSE at all encoder and decoder blocks.
One important aspect of the sSE is it has a sigmoidal gating function at the end before excitation. That means at any location, it has the capacity to saturate all the neurons (i.e. all the output feature map activations becomes 1) which keeps the segmenter feature maps unchanged. Consider such a case where at the encoder/ decoder feature maps are unchanged and just the bottleneck is calibrated. This would be similar to the BL-3 experiment which shows decent performance. Thus, we believe the sigmoidal gating would control the sSE blocks only to re-calibrate the feature maps at scales it is necessary.

\section{Conclusion}
\label{sec:conc}
In this article, we introduced a few-shot segmentation framework for volumetric medical scans. The main challenges were the absence of pre-trained models to start from, and the volumetric nature of the scans. We proposed to use `channel squeeze and spatial excitation' blocks for aiding proper training of our framework from scratch. In addition, we proposed a volumetric segmentation  strategy for segmenting a query volume scan with a support volume scan by strategic by pairing 2D slices appropriately. We evaluated our proposed framework and several baselines on contrast-enhanced CT scans from the Visceral dataset. We compared our sSE based model to the existing approaches based on feature fusion~\cite{rakelly2018few}, classifier regression~\cite{shaban2017one} and their combination. Our framework outperformed all previous approaches by a large margin. 

 Besides comparing to existing methods, we also provided detailed experiments for architectural choices regarding the SE blocks, model complexity, and skip connections. We also investigated the effect on the performance of our few-shot segmentation by changing the support volume and the number of budget slices from a support volume.

Our proposed few-shot segmentation has the following limitations. Firstly, for a new query volume the start and end slices need to be indicated for a target organ to be segmented. This might require manual interaction. Secondly, a very precise segmentation cannot be achieved using few-shot segmentation due to extremely limited supervision and the level of difficulty of this task. If the application demands highly accurate segmentation, we recommend going the traditional supervised learning way by acquiring more annotations for training.

Inspite of the limitations, the exposition of our proposed approach is very generic and can easily be extended to other few-shot segmentation applications. Our approach is independent of pre-trained model, which makes it very useful for non computer vision applications. 

\section*{Acknowledgement}
We thank SAP SE and the Bavarian State Ministry of Education, Science and the Arts in the framework of the Centre Digitisation.Bavaria (ZD.B) for funding and the NVIDIA corporation for GPU donation. 

\section*{References}

\bibliographystyle{model1-num-names}
\bibliography{sample.bib}

\section*{List of IDs in the Visceral Dataset}
The list of IDs in the dataset used for support set, validation query set and testing query set are reported below.

\subsection*{Support Set}
\begin{enumerate}
    \item \texttt{10000132\_1\_CTce\_ThAb}
\end{enumerate}

\subsection*{Validation Query Set}

\begin{enumerate}
    \item \texttt{10000100\_1\_CTce\_ThAb}
    \item \texttt{10000104\_1\_CTce\_ThAb}
    \item \texttt{10000105\_1\_CTce\_ThAb}
    \item \texttt{10000106\_1\_CTce\_ThAb}
    \item \texttt{10000108\_1\_CTce\_ThAb}
    \item \texttt{10000109\_1\_CTce\_ThAb}
    \item \texttt{10000110\_1\_CTce\_ThAb}
    \item \texttt{10000111\_1\_CTce\_ThAb}
    \item \texttt{10000112\_1\_CTce\_ThAb}
    \item \texttt{10000113\_1\_CTce\_ThAb}
    \item \texttt{10000127\_1\_CTce\_ThAb}
    \item \texttt{10000128\_1\_CTce\_ThAb}
    \item \texttt{10000129\_1\_CTce\_ThAb}
    \item \texttt{10000130\_1\_CTce\_ThAb}
\end{enumerate}

\subsection*{Test Query Set}

\begin{enumerate}
    \item \texttt{10000131\_1\_CTce\_ThAb}
    \item \texttt{10000133\_1\_CTce\_ThAb}
    \item \texttt{10000134\_1\_CTce\_ThAb}
    \item \texttt{10000135\_1\_CTce\_ThAb}
    \item \texttt{10000136\_1\_CTce\_ThAb}
\end{enumerate}

\end{document}